\documentclass[conference]{IEEEtran}
\IEEEoverridecommandlockouts
\usepackage{cite}
\usepackage{amsmath,amssymb,amsfonts}
\usepackage{algorithmic}
\usepackage{graphicx}
\usepackage{textcomp}
\usepackage{xcolor}
\usepackage[utf8]{inputenc}
\usepackage{graphicx}
\usepackage{color}

\newcommand{\eg}{{\emph{e.g.}~}}
\newcommand{\ie}{{\emph{i.e.}~}}
\newcommand{\cf}{{\emph{cf.}~}}
\usepackage{subcaption}
\usepackage{balance}

\def\BibTeX{{\rm B\kern-.05em{\sc i\kern-.025em b}\kern-.08em
    T\kern-.1667em\lower.7ex\hbox{E}\kern-.125emX}}
\begin{document}

\title{Customized Video QoE Estimation \\with
Algorithm-Agnostic Transfer Learning\\
}

\author{\IEEEauthorblockN{Selim Ickin}
\IEEEauthorblockA{\textit{Ericsson AB} \\
Stockholm, Sweden \\
selim.ickin@ericsson.com}
\and
\IEEEauthorblockN{Markus Fiedler}
\IEEEauthorblockA{\textit{Blenkinge Institute of Technology} \\
Karlshamn, Sweden \\
markus.fiedler@bth.se}
\and
\IEEEauthorblockN{Konstantinos Vandikas}
\IEEEauthorblockA{\textit{Ericsson AB} \\
Stockholm, Sweden \\
konstantinos.vandikas@ericsson.com}

}

\maketitle

\begin{abstract}
The development of QoE models by means of Machine Learning (ML) is challenging, amongst others due to small-size datasets, lack of diversity in user profiles in the source domain, and too much diversity in the target domains of QoE models. Furthermore, datasets can be hard to share between research entities, as the machine learning models and the collected user data from the user studies may be IPR- or GDPR-sensitive. This makes a decentralized learning-based framework appealing for sharing and aggregating learned knowledge in-between the local models that map the obtained metrics to the user QoE, such as Mean Opinion Scores (MOS). 
In this paper, we present a transfer learning-based ML model training approach, which allows decentralized local models to share generic indicators on MOS to learn a generic base model, and then customize the generic base model further using additional features that are unique to those specific localized (and potentially sensitive) QoE nodes. 
We show that the proposed approach is agnostic to specific ML algorithms, stacked upon each other, as it does not necessitate the collaborating localized nodes to run the same ML algorithm.
Our reproducible results reveal the advantages of stacking various generic and specific models with corresponding weight factors. Moreover, we identify the optimal combination of algorithms and weight factors for the corresponding localized QoE nodes.
\end{abstract}


\section{Introduction}
The ultimate goal for QoE modeling is to develop a model that performs well on  any local QoE dataset, which can be contextually very different. Towards achieving a superior QoE model, a comprehensive user study that covers all possible contexts needs to be performed, which we consider here as a requirement that is hard (if not impossible) to reach, amongst others due to the high effort and cost of the user study. The QoE data in the source domain, where the model is developed, and the QoE data in the target domain, where the model is to be deployed for real operations, need to have a similar distribution. The authors of \cite{wang2018characterizing} show that when a model is trained with features that are only specific to the source domain, \emph{negative transfer} (reducing the model performance)
\begin{figure}
    \centering
    \includegraphics[width=0.9\columnwidth]{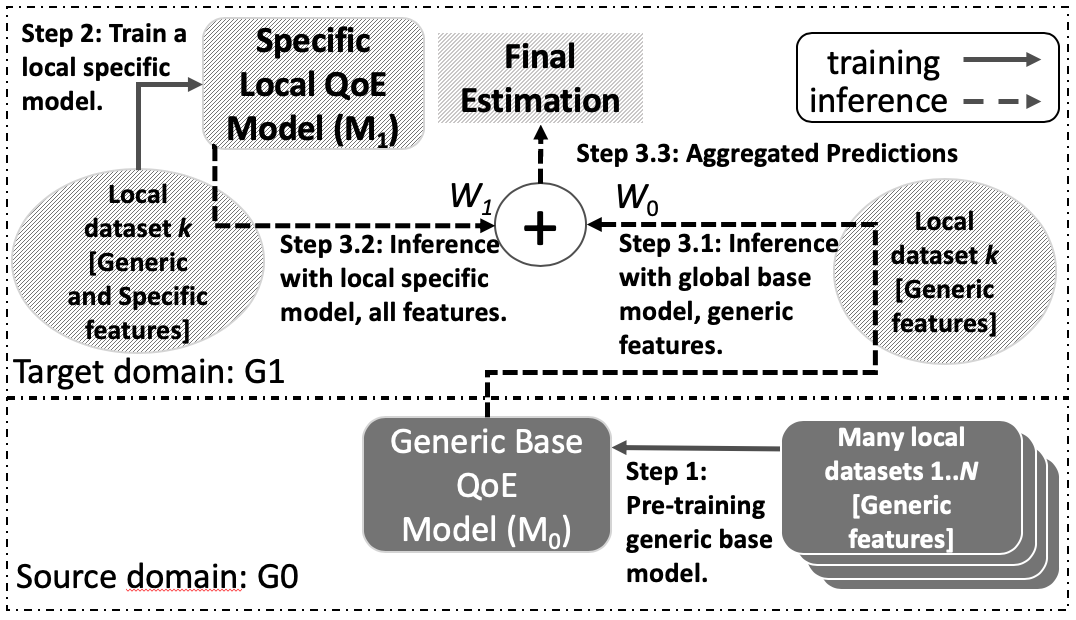}
    \caption{Proposed model training procedure.}
    \label{fig:generic_vs_specific_features}
\end{figure}
can occur, since the specific features at the source domain do not represent the target domain well enough. In parallel, those specific local features in the source domain can be potentially sensitive; detailed user profiles, amongst others, cannot be shared easily \cite{user-privacy-qoe}. Hereby, Machine Learning (ML) \emph{features} refer to a subset of indicative metrics representing the underlying \emph{QoE factors}. In addition, it may happen that one research group is interested in a QoE model particularly for video contents with high spatio-temporal complexity, \eg with the scope of developing action games, while another research group is interested in assessing QoE for a different type of video content.  ML is often preferred in modeling a dataset when the corresponding feature space is large and complex. On the other hand, ML is known as a black-box modelling approach, with frequent lack of \emph{explainability}. In the recent years, human supervision has becoming critical in order to enhance black-box ML models with explainability. 

Our approach consists of both ML and domain expertise, hence we reveal the composite nature of a QoE model that is built on well-defined generic and specific features. We suggest dividing the QoE model into two parts, a \emph{generic base model} that is universally applicable, and a \emph{specific local model} that is specific and dependent on local context. Thereby, we relax the requirements towards the goal of achieving one superior customized QoE model. We hypothesize that, when a generic base model is used, a QoE model customization at the target domain is needed to improve the accuracy further, since the generic model in the source domain lacks additional features that are strongly representative for the target domain. Customization of the pre-trained model is done via stacking the pre-trained model with the one trained at the local node. Stacking, which is a combination of  predictions of different independent algorithms, is known to perform better than each algorithm alone\,\cite{stacked-generalization}. This process does not necessitate the two models being of the same algorithm type, hence making the process \emph{agnostic} to the involved ML algorithms, which provides the freedom of choice of algorithms that fit best to global and local domains, respectively. Thus, in this paper, we present an algorithm-agnostic transfer learning method and demonstrate a scenario consisting of i) training a generic QoE model with generic features at the source domain, ii) training a separate specific local QoE model with all features (together with the ones that are specific to target domain) at the target domain, and iii) stacking the aforementioned two models as sketched in Figure~\ref{fig:generic_vs_specific_features}. This way, the local QoE model at the target domain can further be improved while benefiting from the pre-trained base generic model, potentially trained on a rather large dataset. \par

\par
In classical QoE studies, there are (rather strict) assumptions about which context and content parameters matter.  Hence, an additional process of distinguishing the generic features from the specific ones needs to be developed. In QoE studies, there is a wide variety of content being shown to users for assessments, where the dataset includes some features that are only specific to that group. 
Since a consensus of running large scale QoE experiments exactly on same or similar video content is not feasible (if ever possible), it could be a good practice to distinguish those features from the training of the generic base model. This way, the generic base model would be trained only with generic features that yield similar effects on the MOS values that are ideally universally applicable. In the scope of this paper, we choose  features related to playback interruption and playback resolution quality for the purposes of training ML models. 

This paper is structured as follows. We present a detailed explanation of the proposed algorithm-agnostic transfer learning method for video QoE model development in Section \ref{procedure-overview}. In Section \ref{related-work}, we present related work on transfer learning and its challenges. The algorithms, model performance evaluation, and the dataset used in our experiments are presented in Sections \ref{sec:algorithms} and \ref{dataset-feature-extraction}, respectively. We motivate the selection of video content-based specific features via various data-driven approaches in Section \ref{sec:effectofvideo}. We split the dataset into source and target domain and prepare them for the transfer-learning experiments. Results are given in Section \ref{sec:results} where we show the advantages of stacking generic and specific local model with varying weight factors. The paper is concluded in Section \ref{sec:conclusion}, accompanied by a brief discussion on limitations and directions for future work. 

\section{Procedure Overview}\label{procedure-overview}
The procedure consists of multiple steps as described in Figure \ref{fig:generic_vs_specific_features}. The assumption is that there exists a base model, which is pre-trained with a rather large dataset via participation of preferably multiple research entities (G0 at source domain) using generic features as shown in Step 1. In Step 2, a new research group (G1 at a target domain) trains a separate ML model with any choice of feature set locally unique to the collected data samples in the target domain. In Step 3, the research group (G1 at target domain),  transfers the pre-trained generic  model together with the metadata containing information about the set of generic features. Next, the locally trained specific model and the transferred generic model are stacked together and this customized fine-tuned model is deployed locally at G1. The stacking is a weighted average of the predictions of the base model and the local model on the testset as given in  Eq.\,\ref{eq:stacking}.
\begin{equation}
{y'}_{\textrm{test},\textrm{G1}} = W_{0}  M_{0,GF}(X_{\textrm{test},\textrm{G1}}) + W_{1}  M_{1,[GF,SF]}(X_{\textrm{test},\textrm{G1}})
\label{eq:stacking}
\end{equation}
$M_{i,j}$ is the trained model with the training set of group $i$, using the feature group $j$. For the transferred $M_0$, the feature group consists of generic features (GF), while both GF and specific features (SF) are present during the training of $M_1$. The model weights $W_i$ for group $i$ are decimal numbers within the range of 0 and 1, which add up to 1. ${y'}_\textrm{test, G1}$ is the weighted average of the two trained models' inference on the test set of G1, $X_{\textrm{test}, \textrm{G1}}$. 
\section{Related Work}\label{related-work}
Decentralized learning techniques such as transfer learning and federated learning are well-known\cite{transfer-learning-survey-paper}, however we have not seen many example applications within  the QoE domain, which we believe could be very suitable. In \cite{emotion-transferlearning}, the authors used transfer learning to estimate the labels of an unlabeled dataset where the labels represent user emotions. In \cite{IVF-19}, possible collaborative decentralized learning techniques are presented within the scope of web QoE, which necessitates a neural network algorithm on the contributing local nodes. In this paper, we present an algorithm-agnostic transfer learning based method, which in principle does not limit the algorithm to a neural network. 

In decentralized learning, \eg federated learning, models are trained collaboratively by combining different models which have been trained locally, to a single model. When such a model is shared back to the individuals that contributed to training, immediate benefits can be detected for those that have similar data distribution. In \cite{2019federated}, feature selection is studied within neural network based federated learning, where the model parameters are divided into private and federated parameters. Only the federated parameters are shared and aggregated during federated learning. It is shown that this model customization  approach significantly improves the model performance while preserving privacy.

From a privacy perspective, techniques such as differential privacy\,\cite{dwork} and secure aggregation\cite{bonawitz} can be utilized for sharing private information for the purposes of training ML models without revealing the original dataset i.e. via attacks such as Deep Leekage\cite{deep_leekage}. Even though these techniques work well in the cases where information needs to be exchanged, in the scope of this work they are not applicable since we only share the generic model (architecture and internal representation) which is already known and as such does not need to be protected.

\section{Algorithms and Evaluation}\label{sec:algorithms}
We used XGBoost, a boosting algorithm, available within the Python Scikit-learn library, in the experiments as it is known for its superior overall performance in tabular datasets. XGBoost is powerful, because it is an ensemble of different decision trees, where predictions of each tree are averaged together to give out a final decision. Split boundaries in the trees are obtained using an iterative process involving first and second derivatives of the loss function w.r.t. prediction evaluated at each data sample. 
It has also trainable properties, that allows the local workers to update the model with generic features and contribute to the generic model after usage. We set the hyper-parameters as given in the first row in Table\,\ref{tab:model_hyperparameters}. 

\begin{table}[h!]
    \centering
    \begin{tabular}{|c||c|}
        \hline
        XGBoost\cite{chen2016xgboost} & eta: 0.004; max depth: 4; subsample: 0.5; \\
        (Python)& colsample bytree: 1; objective: reg:squarederror;\\
        Scikit-learn& eval metric: rmse.   \\
        \hline
        NN (Python) & Layer 1: 32, Layer2: 64 neurons; eval. metric: mse;  \\
        Tensorflow/Keras& learning rate: 0.001; activation: relu.; dropout: 0.3.\\
        \hline
        M5P \cite{WW-97} (Weka) & model tree, min. number of instances per leaf: 4 \\
        \hline
        \end{tabular}
    \caption{Model hyperparameters.}
    \label{tab:model_hyperparameters}
\end{table}

\par
In addition, we included a neural network in our experiments to test its applicability in the cases where a base model is pre-trained via federated learning. The Neural Network (NN) algorithm consists of neurons where their goal is to tune the weights of the connection (\ie applying a non-linear function on the multiplication of each input signal to each neuron with a corresponding weight) in between the neurons such that the loss between the predictions of the neurons at the last layer and the target variable is minimized. We used the hyper-parameters as given in the second row of Table\,\ref{tab:model_hyperparameters}.\par
We perform over 100 experiments to achieve statistical significant comparison of scenarios. In each experiment, we randomly select $70\,\%$ of data as training-, and $30\%$ as test-set. We use the correlation coefficient $R^2$ and MAE (Mean Absolute Error), indicating the mean absolute difference between the truth label and the predictions, as accuracy evaluation metrics. 95\,\% confidence interval half-sizes are given in brackets.
\par
In a complementary approach, we train a piece-wise decision tree model with the \emph{M5P} ML algorithm \cite{WW-97} to understand the effect of content features in an instructive explainable \emph{model tree}.
The latter is a representation of a set of QoE models with (usually) several ML features as input parameters, each valid over a subset of the feature space, where the subsets are defined by a decision tree. In earlier works \cite{CW-17, FCP-19}, M5P has shown its ability to identify a set of well-explainable QoE models. As our primary goal is to visualise and explain, over-fitting is not an issue, and thus, we train and test on the full data set. In the sequel, we use the Weka \cite{EHW-16} 3.8.4 implementation of M5P with the default settings listed in the last row of Table\,\ref{tab:model_hyperparameters}.
 
\section{Dataset and Feature Extraction} \label{dataset-feature-extraction}
The publicly available QoE dataset (Waterloo Streaming QoE Database III (SQoE-III)) \cite{duanmu2018sqoe} is used in this paper. It is collected using a well-grounded user study methodology. The MOS scores on the dataset are highly correlated (higher than 0.8) with the latest published QoE models. The MOS scores also had a Pearson correlation of at least 0.6 with six existing QoE models.
The dataset consists of 20 raw HD reference DASH videos, where the video length is 13\,s on average. The experiments are performed while changing the video streaming bandwidth amongst 13 different bitrate levels ranging from 0.2\,Mbps to 7.2\,Mbps. The switch is performed in six categories; stable, ramp-up, ramp-down, step-up, step-down, fluctuation. The video sequences vary in spatio-temporal complexity.  
There were 34 subjects (ages range between 18 and 35) involved in the study (where 4 turned out to be outliers, hence they are removed from the study). The user ratings are collected using a rating scale, where the scores are between 0 (worst) and 100 (best). 

We extracted the features as described in Table \ref{tab:feature_desc} from the raw dataset. The feature names are also depicted with a label indicating whether or not the feature is selected to be a generic or specific feature.
\begin{table*}[h!]
    \centering
    \begin{tabular}{|c|c||c|c|}
        \hline
        feature  & description & feature  &description \\
        \hline\hline
        TI (SF) & Video temporal complexity   & stallTimeInitialTotal (GF)& Initial stalling/buffering duration    \\
        \hline
        SI (SF)& Video spatial complexity   & meanBitrate (GF)& Average playout bitrate \\
        \hline
        fps (GF)& Played frames per second  & bitrateTrend (GF)& Slope of the video playout bitrate  \\
        \hline
        nstalls (GF)& Number of stalls & lastbitrate (GF)& Last playout bitrate  \\
        \hline
        stallTimeIntermediateTotal (GF)& Total stalling duration & \textbf{MOS} (target variable) & Mean Opinion Score \\
        \hline\end{tabular}
    \caption{Descriptions of the extracted features for the QoE models, where the Mean  Opinion Score (MOS) is the target variable of the model.}
    \label{tab:feature_desc}
\end{table*}
\begin{itemize}
    \item \textbf{\emph{Generic features (GF):}}
We define those as (QoE) factors that have similar effect on human perception. For example, increase in playout bitrate increases the QoE of users, since videos that are presented at a high resolution are perceived better as compared to lower resolutions. Also, stalling events in a video streaming have a negative impact on video QoE \cite{stalling}.
    \item \textbf{\emph{Specific features (SF):}} 
Those are the features that are merely specific to the local dataset where the model is deployed at. For instance, it can be related to the specific characteristic of the content, the user groups, the geographic region or other particularities of the model.
\end{itemize}

\section{Effect of Video Content on QoE} \label{sec:effectofvideo}
In this paper, we hypothesise through domain expertise that the specific features are the ones related to the spatio-temporal content complexity, represented by the TI (Temporal Information) and SI (Spatial Information) features. In order to quantify the effect of the video content on the QoE modeling, we perform a set of tests.

\subsection{Training With or Without Content Features}
We first train a ML model both with and then without the specific TI and SI features. Next, the model accuracy performance of each case is compared. In Table \ref{tab:contentdependencyoverall}, the performance from both scenarios (with or without the specific features) is depicted. ``Without content'' means that TI and SI features are omitted; ``With content'' means that TI and SI features are included. We observe that the inclusion of content related features makes the model a more powerful predictor (with a rather high $R^2$ and low MAE), indicating that content-related features are important QoE factors.
\begin{table}[h!]
    \centering
    \begin{tabular}{|c||c|c|}
        \hline
        Evaluation Metric & without content & with content \\
        \hline\hline
        $R^2$ &  0.74(0.01) & 0.80(0.01) \\
        \hline
        MAE & 6.16(0.07)& 5.41(0.07)\\
        \hline\end{tabular}
    \caption{XGBoost ML model performance on all dataset with and without the content specific features. }
    \label{tab:contentdependencyoverall}
\end{table}
\subsection{SHAP Sensitivity Analysis} \label{sec:SHAP}
\begin{figure}[h]
    \centering
        \includegraphics[width=0.85\columnwidth]{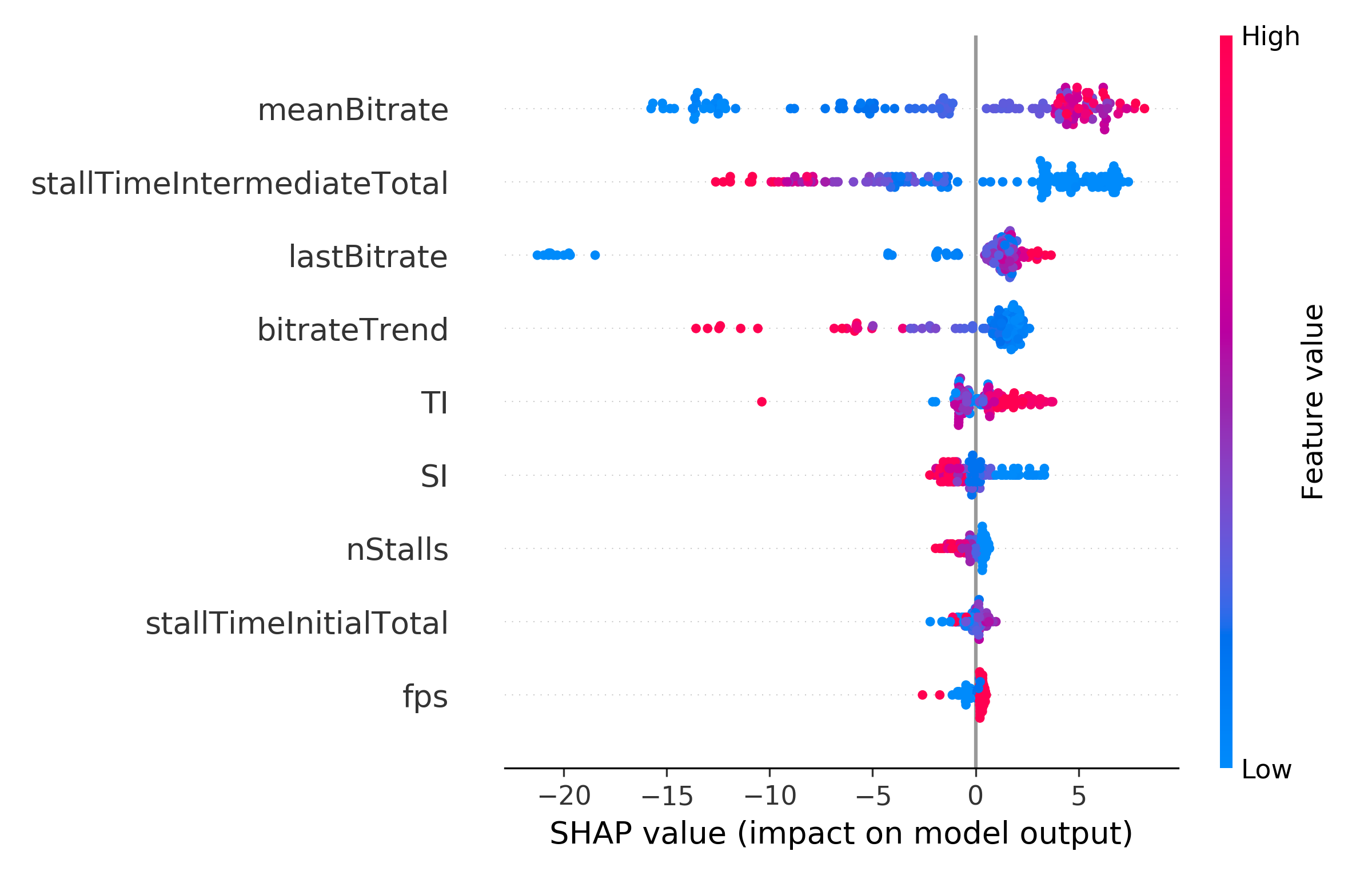}
        \caption{SHAP values of input features.} 
        \label{fig:SHAPoverall}
    \end{figure}

We use the TreeExplainer by SHAP\cite{NIPS2017_7062} to apply sensitivity analysis and observe the features that are important for the decision of the trained ML model. In Figure \ref{fig:SHAPoverall}, the effect of each input feature on model prediction is quantified from most to least important (descending order from top to bottom on the y-axis). Red regions indicate that the absolute value of the effect of the feature on the model prediction is high, while blue regions indicate the contrary. The dots being on the positive side of y-axis indicate that the effect is positive, a higher feature value contribute to a higher MOS, and vice-versa if they are located at the negative side. The SHAP diagram indicates that the average bitrate (meanBitrate) is the most important metric in the model prediction, contributing to raising the MOS value as expected. Similarly, the total video stall duration (stallTimeIntermediateTotal) has a high influence on the negative side, which implies that it tends to lower the expected MOS value. The TI and SI features are considered more important than other features such as the number of stalls (nStalls) and the initial stalling time (stallTimeInitialTotal). 

We conclude that TI and SI are important features to consider, and we create and experiment a scenario by splitting the whole dataset into two groups and develop two separate QoE models based on different content types. 
\subsection{Content-Based Split}
\begin{figure*}[h!]
    \centering
    \includegraphics[width=\textwidth]{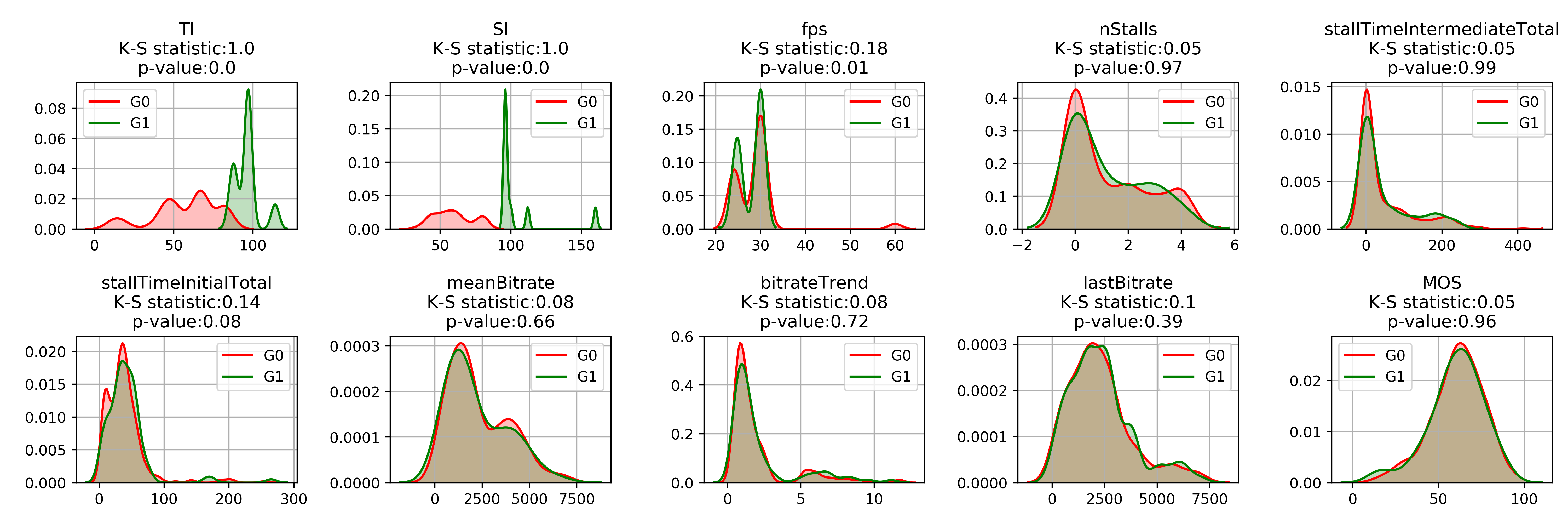}
    \caption{Kernel Density Estimate plots of all features in G0 and G1 is given.}
    \label{fig:feature_dist_groups}
\end{figure*}

Our choice of thresholds $\mathit{SI} = 85$ and $\mathit{TI}  = 85$ leads to two groups. Data samples with TI and SI feature values smaller than 85 are assigned to G0 (the source domain), and the remaining data samples are assigned to G1 (the target domain). The dataset sizes are 353 and 97 for groups G0 and G1, respectively. We selected G0, with a high number of data samples, to train a generic transferable QoE model, and a smaller group (G1) to represent a small QoE dataset that can potentially benefit from the generic model that is trained on the G0 dataset. 
The distribution of all features on the two groups is illustrated in Figure \ref{fig:feature_dist_groups} together with the Kolgomorov-Smirnov test results. Having a low K-S statistics with a high p-value for a feature indicates no statistically significant difference in the probability distribution between the two datasets, at 99\,\% confidence. We observe this on all features except of TI and SI. It can be further argued that fps is likely correlated with the content features, hence manifesting a statistically different distribution if the confidence interval was 95\%. Thus, in our splitting process, we choose TI and SI values as the only specific local features in this paper.    

\subsection{Random Split}
In order to quantify the (presumably advantageous) effect of content-based features, we shuffle the entire dataset in two groups. In such a random split, there is a homogeneous distribution of each content in each group.
Groups G0 and G1 have 353 and 97 random samples, respectively, \ie the same data sample size as in the content-based split.\par 
In Table\,\ref{tab:randomvscontentsplit}, the performance of the two models is compared, using the $R^2$ evaluation metric. Obviously, when groups have homogeneous samples from different content types, the model becomes less accurate. However, when trained on samples that are from the same content category, the accuracy on G0 increases from 0.77 to 0.80. The gain is even more visible on G1 where the accuracy increases from 0.61 to 0.73. Thus, we find our presumption that content splitting improves modeling performance confirmed.
\begin{table}
    \centering
    \begin{tabular}{|c||rc|rc|}
        \hline
        Evaluation Metric & \multicolumn{2}{c|}{random split} & \multicolumn{2}{c|}{content (TI/SI) split} \\
        \hline\hline
            & Overall: & 0.69(0.02) & Overall: & 0.77(0.01)  \\
    $R^2$   & G0: & 0.77(0.01) & G0: & 0.80(0.01)  \\
           & G1: & 0.61(0.02) & G1: & 0.73(0.02)\\ 
        \hline\end{tabular}
    \caption{ML model performance on different splits. }
    \label{tab:randomvscontentsplit}
\end{table}

\subsection{M5P Model Trees}


In the M5P experiments, we focus on the differences made by the inclusion of content on model trees and matching performances.
We confine ourselves to training and testing on the same group, and describe the properties of the emanating trees, which had to be omitted due to space limitations.

In case of G0, the majority of the available features are included in the decisions. Including the content makes the model tree grow from 5 to 7 and the model correlation grow from $R^2 = 0.90$ to 0.92. Feature SI appears in the decisions close to some few leaves, while feature TI is omitted from the decisions. The absolute weights of the SI feature are up to two orders of magnitude larger than the ones belonging to the TI feature.

In case of G1, the decision trees are compact (3 leaves), with only the feature meanBitrate involved (\cf Section\,\ref{sec:SHAP} regarding its importance). Including the content features, makes the correlation grow from $R^2 = 0.89$ to 0.95. Both TI and SI appear in the linear models, where the weight factors of TI are one order of magnitude larger than those of SI.

Obviously, the model trees have rather different characteristics, which reinforces the view that they complement each other. Integrating both groups in the same model tree implies a more complex structure (7 resp. 9 leaves) at performance comparable to G0 ($R^2 = 0.90$ resp. 0.91), where the SI and TI features become present in both decisions and linear models as soon as they get included, indicating their key role in the modeling process. 


\section{Results}\label{sec:results}

Table \ref{table:crossperformance} summarises results of our experiments. Column~1 is the group id where the models are trained on. Columns~2--5 contain the accuracy performance of those trained models on the testset of different groups. 
The table splits the results based on the scenarios whether or not they are \emph{trained} using the content-based SF. 
Not surprisingly, higher individual model performance is observed when models are trained with content-related features. The model accuracy increases from 0.75 to 0.80 in G0, and from 0.56 to 0.73 in G1. 
In the latter case, training in isolation with a highly specific local dataset yields a significantly better performance.
However, the question remains how G1 can benefit from G0.


\begin{table}[h!]
\begin{center}
\begin{tabular}{|c||c|c|c|c|}
\hline
 & \multicolumn{4}{c|}{\textbf{Test On}}  \\
 & \multicolumn{2}{c|}{\textbf{Without Content}}&\multicolumn{2}{c|}{\textbf{With Content}}  \\
\textbf{Train On}  & G0 & G1 & G0 & G1\\
 \hline
 \hline
G0 & 0.75(0.01) & \textbf{0.71(0.01)} &0.80(0.01)  &\textbf{0.68(0.02)}  \\
\hline
G1 & N/A &0.56(0.03) & N/A  &\textbf{0.73(0.01)}  \\
\hline



\end{tabular}
\end{center}
\caption{Local and cross performance of the models.}
\label{table:crossperformance}

\end{table}

\subsection{Cross-Group Performance}

We focus on model transfer from a source domain having a representative group with large data samples such as G0, to a target domain with small sample size such as G1. We observe the cross performance of the generic base model $M_0$ on G1, in order to observe the effect of negative transfer due to SF.  We perform this test by transferring $M_{0,GF}$, and then transferring  $M_{0,[GF,SF]}$. The model $M_{0,GF}$ performs with an accuracy of 0.75 in G0, while performing with accuracy of 0.71 on G1, with a decrease of 0.04 in accuracy. $M_{0,[GF,SF]}$ performs  0.80 on G0, while performing 0.68 on G1 with a decrease of 0.12 in accuracy. When $M_{0,[GF,SF]}$ is transferred, the decrease in accuracy is higher (0.12 instead of 0.04). This is due to the fact that content specific features at the source domain are important, but different in the target domain.

\subsection{Stacked Model Using Transfer Learning}
\begin{figure}[h!]
    \centering
    \includegraphics[width=0.9\columnwidth]{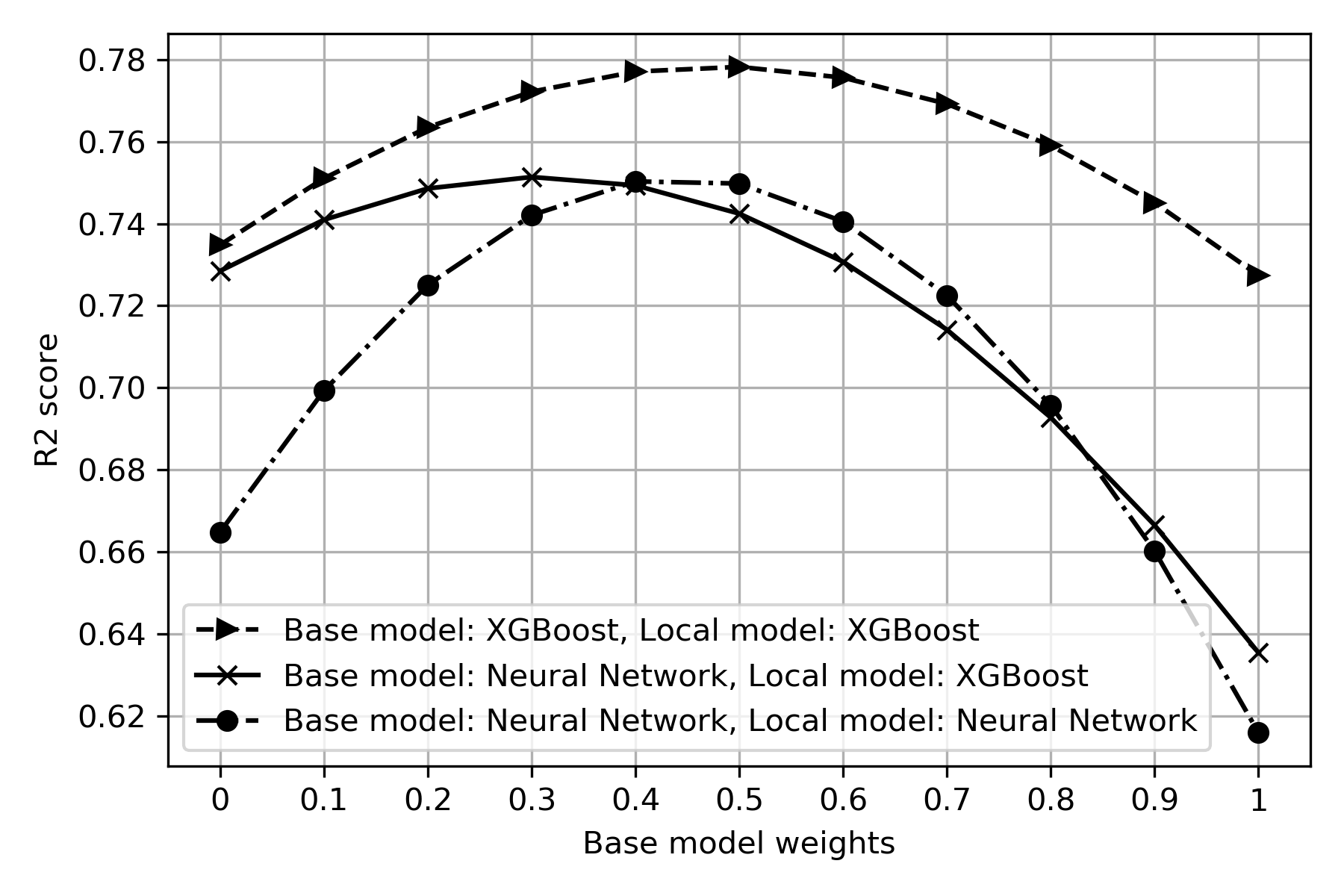}
    \caption{Performance of the stacked models for two different generic base model algorithms with varying corresponding base model weights ($W_0$).}
    \label{fig:scan_basemodel_weights}
\end{figure}

In order to achieve a more accurate model on G1, we combine $M_{1,[GF,SF]}$ with the generic learnings from $M_{0,GF}$, exploiting the algorithm-agnostic nature of our approach. We tested how this stacked model performance (according to Equation \ref{eq:stacking}) changes with respect to varying weights. When the same algorithms (either XGBoost or Neural Network) are deployed at source and target domain, a weighting factor of base model $W_0 = 0.5$ yielded the best accuracy, as shown in Figure\,\ref{fig:scan_basemodel_weights} and summarised in Table\,\ref{tab:stacked_model_comparison}. The accuracy of the model in G1 further increased from 0.73 (stand-alone local) to 0.78 (stacked local) when the XGBoost model is used at both ends. This indicates that involvement of target and source domain models equally increases the model performance to the greatest extent. We also tested a scenario when $M_{0,GF}$ is a Neural Network, and $M_{1,[GF,SF]}$ is an XGBoost. This scenario may occur when the base model is a pre-trained Neural Network with a federated learning mechanism as in \cite{IVF-19}. In this case, we again see the benefit of stacking at $W_0=0.3$, which indicates that the generic base model is a rather weak contributor. We presume that the Neural Network model could not capture the data pattern in G0 as good as XGBoost did. This indication is also inline with poor (just below $0.62$) accuracy when the weight of the Neural Network base model $M_{0,GF}$ is 1, i.e. the local model is excluded. 
\begin{table}[]
    \centering
    \begin{tabular}{|cc||c|c|}
\hline
\multicolumn{2}{|c||}{Scenario} & & $M_{0,GF}$ \\
Base ($M_{0,GF}$) & Local($M_{1,[GF,SF]}$) & $R^2$ & Optimal $W_0$ \\
\hline \hline
NN & NN   & 0.75(0.01) &0.5  \\
\hline 
NN & XGBoost  & 0.75(0.01)&0.3  \\
\hline 
XGBoost & XGBoost &\textbf{0.78(0.01)} &0.5 \\

\hline 
\end{tabular}
\caption{Comparison of accuracy values for different algorithm combinations.}
\label{tab:stacked_model_comparison}
\end{table}

\balance 

\section{Conclusion}\label{sec:conclusion}
In this paper, we first reveal the composite nature of a QoE model, by decoupling the specific QoE factors from the generic ones, using both domain-expertise and data-driven approaches. This enables the remote nodes to share knowledge without having to share sensitive local specific information. We demonstrate the knowledge sharing via transfer learning followed by stacking process. We present that this process in parallel helps to customize the local model with the help from the generic model, in the case when there is no sufficient local training dataset. In addition, due to the nature of stacking method, the models that participate in the training can be of any algorithm, hence can freely be selected. We show that a small QoE dataset with specific features can benefit (with a $7\%$ improvement in estimation accuracy) further from a generic model via our privacy-preserving machine learning setup.  \par 
While performing this study, certain limitations came to our attention. The video sequences used in the QoE study are short, hence the features in the generic and specific categories might be slightly different on longer video tests. On the other hand, since our method is data-driven, the generic and specific features can be re-categorized based on other datasets. The feature set size that we extracted from the raw data is limited to 9, however we believe that the extracted features are sufficient to represent a video QoE model and to communicate our purpose in the paper. \par 
In future work, we plan to derive more features both for generic and specific categories, while not limiting the specific features to content-type but also other context features, amongst others related to user profile, experiment setting details and device type, respectively.

  \bibliographystyle{abbrv}
  \bibliography{bibliography}

\end{document}